\documentclass[conference, hidelinks]{IEEEtran}
\IEEEoverridecommandlockouts

\usepackage[utf8]{inputenc}
\usepackage[T1]{fontenc}
\usepackage{orcidlink}
\usepackage{cite}
\usepackage{amsmath,amssymb,amsfonts}
\usepackage{algorithmic} 
\usepackage{graphicx}
\usepackage{textcomp}
\usepackage{xcolor}

\def\BibTeX{{\rm B\kern-.05em{\sc i\kern-.025em b}\kern-.08em
    T\kern-.1667em\lower.7ex\hbox{E}\kern-.125emX}}

\begin{document}

\title{%
  {\small
    \textbf{PREPRINT}: Copyright IEEE 2025. This is the author's version of the work. It is posted here for your personal use. Not for redistribution.\vspace{-2em}
    \\
    The definitive version is accepted and published by The 20th Annual System of Systems Engineering Conference (SOSE25).
  }\\[0.7em]
  Urban Air Mobility as a System of Systems: An LLM-Enhanced Holonic Approach
}

\author{\IEEEauthorblockN{Ahmed R. Sadik}
\IEEEauthorblockA{\textit{Senior Scientist} \\
\textit{Honda Research Institute Europe}\\
Offenbach am Main, Germany \\
ahmed.sadik@honda-ri.de~\orcidlink{0000-0001-8291-2211}}
\and
\IEEEauthorblockN{Muhammad Ashfaq}
\IEEEauthorblockA{\textit{Faculty of IT} \\
\textit{University of Jyväskylä}\\
Jyväskylä, Finland \\
muhammad.m.ashfaq@jyu.fi~\orcidlink{0000-0003-1870-7680}}
\and
\IEEEauthorblockN{Niko M\"{a}kitalo}
\IEEEauthorblockA{\textit{Faculty of IT} \\
\textit{University of Jyväskylä}\\
Jyväskylä, Finland \\
niko.k.makitalo@jyu.fi~\orcidlink{0000-0002-7994-3700}}
\and
\IEEEauthorblockN{Tommi Mikkonen}
\IEEEauthorblockA{\textit{Faculty of IT} \\
\textit{University of Jyväskylä}\\
Jyväskylä, Finland \\
tommi.j.mikkonen@jyu.fi~\orcidlink{0000-0002-8540-9918}}
}

\maketitle

\begin{abstract}

Urban Air Mobility (UAM) is an emerging System of System (SoS) that faces challenges in system architecture, planning, task management, and execution. 
Traditional architectural approaches struggle with scalability, adaptability, and seamless resource integration within dynamic and complex environments.
This paper presents an intelligent holonic architecture that incorporates Large Language Model (LLM) to manage the complexities of UAM. Holons function semi-autonomously, allowing for real-time coordination among air taxis, ground transport, and vertiports. LLMs process natural language inputs, generate adaptive plans, and manage disruptions such as weather changes or airspace closures.
Through a case study of multimodal transportation with electric scooters and air taxis, we demonstrate how this architecture enables dynamic resource allocation, real-time replanning, and autonomous adaptation without centralized control, creating more resilient and efficient urban transportation networks.
By advancing decentralized control and AI-driven adaptability, this work lays the groundwork for resilient, human-centric UAM ecosystems, with future efforts targeting hybrid AI integration and real-world validation.

\end{abstract}


\section{Introduction}
\label{sec:introduction}


Urban Air Mobility~(UAM) represents a transformative paradigm that integrates autonomous aerial and ground systems into a single, unified, multimodal transportation framework~\cite{10388419, xiang2024autonomous}. 
As cities worldwide face unprecedented congestion challenges---with commuters in megacities spending significant time daily in traffic~\cite{koreaHerald} and urban populations projected to increase by 2.5 billion by 2050~\cite{UNUrbanizaiton}---traditional transport infrastructure is reaching its physical and operational limits. 
UAM promises to alleviate these pressing challenges by utilizing the underexploited vertical dimension of urban spaces to reduce travel times by up to 30--50\% compared to ground transportation in congested areas~\cite{moradi2024urban}.

\begin{figure}[!t]
\centering
\includegraphics[width=.9\linewidth]{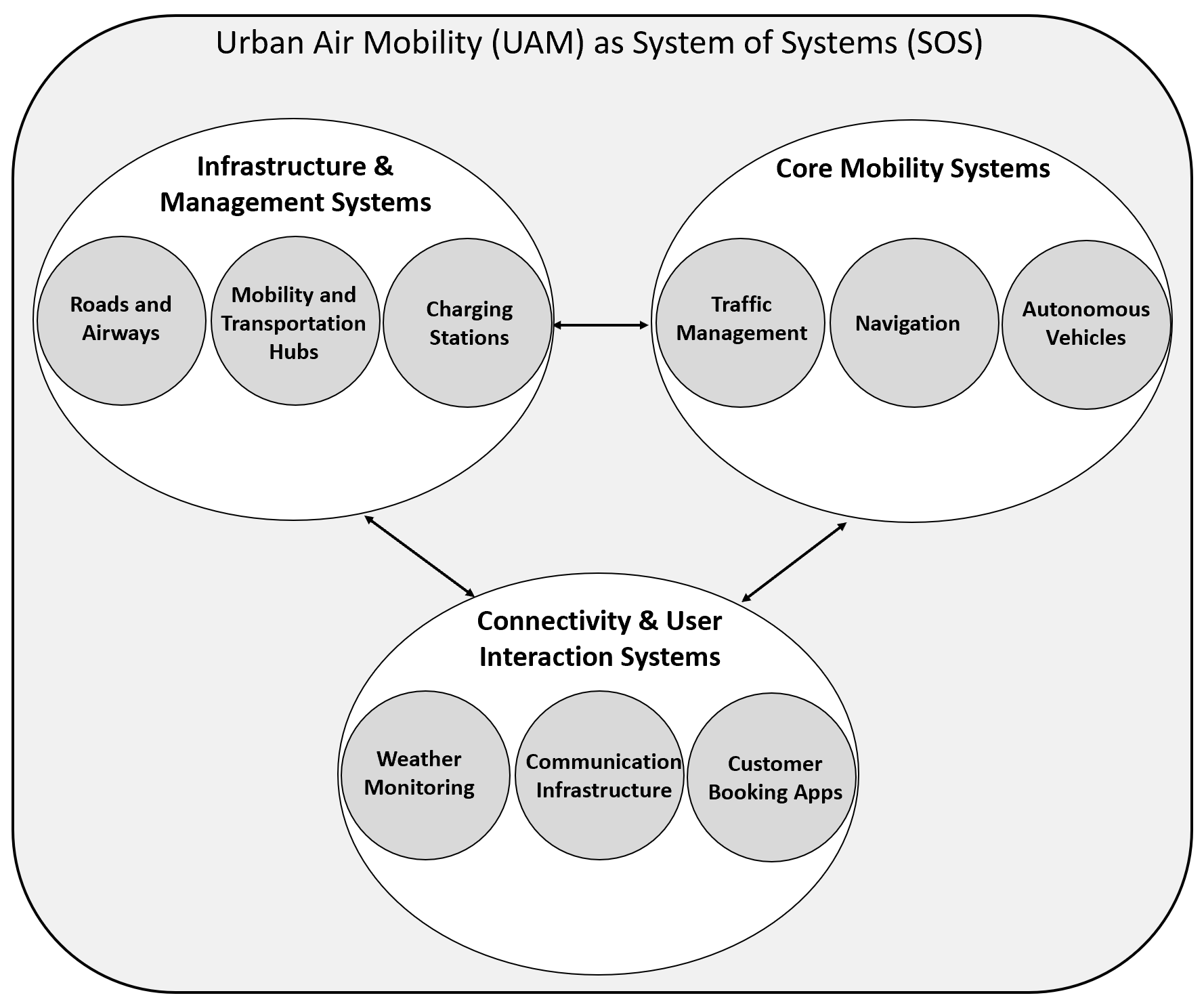}
\caption{Urban Air Mobility as System of Systems.}
\label{fig:UAM_SOS}
\end{figure}

Conceptualized as a System of Systems (SoS), UAM comprises independent yet interdependent subsystems---including air taxis, ground transport, and air traffic control---that collaboratively enhance efficiency, scalability, and safety~\cite{rubio2023urban} (see Fig.~\ref{fig:UAM_SOS}).
Unlike conventional systems built on centralized control and predictable interactions, an SoS integrates independent systems whose dynamic interactions produce emergent behaviors~\cite{maier1998architecting, boardman2006system}. 

The architectural requirements of UAM are fundamentally shaped by its SoS nature~\cite{sadik2023self,brulin2025system}. Table~\ref{tab:sos_comparison} contrasts traditional systems with SoS, highlighting key differences in autonomy, interoperability, diversity, and emergent behaviors~\cite{ISO2019}. Unlike monolithic networks managed by central authorities, UAM demands decentralized mechanisms for effective operation and coordination.

In UAM, subsystems operate independently but must interoperate seamlessly to sustain a cohesive mobility ecosystem. This decentralized structure introduces several challenges:

\begin{itemize}
    \item Coordination Complexity: Managing hundreds of aerial vehicles in constrained urban airspace while accounting for dynamic factors such as weather disruptions~\cite{fontaine2023urban, sengupta2023urban}.
    \item Intermodal Integration: Achieving seamless transitions between ground and air transport modes and addressing capacity limitations at vertiports~\cite{fontaine2023urban}.
    \item Interoperability: Implementing network-centric approaches that enable dynamic communication among diverse platforms, including drones, autonomous vehicles, smart city infrastructures, and user-facing applications~\cite{Henshaw2013, DANSE}.
    \item Emergent Behaviors: Managing unpredictable system phenomena, such as demand-driven resource allocation, adaptive route planning, and congestion resolution, that cannot be anticipated or controlled by centralized mechanisms~\cite{fontaine2023urban}.
\end{itemize}

These complexities highlight several critical architectural requirements: First, the system architecture must be modular, interoperable, and capable of dynamic evolution while incorporating real-time coordination and adaptive decision-making capabilities. Operational control must effectively coordinate independent yet interdependent entities without relying on a single controlling authority. Resource allocation requires integrated management of energy, airspace, and fleet resources across diverse systems. Finally, security and safety demand robust communication protocols, resilience mechanisms, and recovery strategies in uncertain environments, necessitating a shift toward distributed architectures.

Recent research highlights the holonic architecture as a particularly promising approach for managing complexity in distributed systems~\cite{elhabbash2024principled}. 
A holonic architecture organizes systems into \textit{holons}---semi-autonomous, recursively nested units that inherently function as both self-contained entities and interdependent components within a hierarchical structure~\cite{blair2015holons}.
This structure enables real-time adaptability, self-organization, and efficient coordination across dynamic, decentralized environments~\cite{indriago2016h2cm}.

In the context of UAM, the holonic architecture allows components such as air vehicles, ground transport, and vertiports to operate independently while still forming a cohesive, resilient transportation network capable of adjusting to changing conditions without centralized control mechanisms.

While traditional AI methods (e.g., rule-based engines or narrow supervised models) excel at well-defined tasks, they struggle with unstructured inputs and open-ended reasoning~\cite{alahakoon2023self}. 
Large Language Models (LLMs) address these challenges by interpreting natural-language commands and dynamically adapting plans to novel scenarios (e.g., weather disruptions, airspace closures)~\cite{Sha2024Generative}.

Building upon our earlier work~\cite{ashfaq2025llm}, which introduced a general LLM-driven holonic architecture for SoS, this paper adapts it to the unique demands of UAM, including multimodal transit and no-fly zones. We condense the core architectural concepts and tailor them to UAM-specific components and domain-specific constraints, demonstrating the approach through a detailed case study that coordinates both ground and air mobility.

The remainder of the paper is organized as follows.
Section~\ref{sec:background} reviews existing federated architectures in UAM and introduces the LLM-enhanced architecture.
Section~\ref{sec:case-study} presents a UAM-specific case study that demonstrates the benefits of the proposed architecture. It also describes the decomposition of the UAM-SoS into semi-autonomous holons capable of dynamically adapting to both local and global requirements.
Section~\ref{sec:discussion} discusses the broader implications and key advantages of the proposed approach.
Section~\ref{sec:conclusion} concludes the paper and outlines directions for future research.

\begin{table}[t]
\caption{Comparison between a System and a System of Systems}
\begin{center}
\begin{tabular}{|p{1.8cm}|p{2.8cm}|p{2.8cm}|}
\hline
\textbf{Characteristic} & \textbf{System} & \textbf{System of Systems} \\
\hline
\textbf{Autonomy} & 
Conformance: The system is autonomous by design, as components cede the autonomy. & 
Independence: SoS is autonomous by integration, as the constituent systems (CS) freely exercise autonomy. \\
\hline
\textbf{Belonging} & 
Centralization: Parent-child relation is the main link among the system components. & 
Decentralization: Systems choose to belong on a cost/benefit basis, optimizing their purposes. \\
\hline
\textbf{Interoperability} & 
Platform-centric: System components are designed to achieve a specific purpose through predictable communication. & 
Network-centric: Dynamic communication in an ad hoc style enhances CS's capabilities. \\
\hline
\textbf{Diversity } & 
Homogeneous: The system's hardware and software are often homogeneous. & 
Heterogeneous: SoS contains a variety of software and hardware that interoperate. \\
\hline
\textbf{Emergence} & 
Foreseen: The system behaviour is static and foreseen by design. & 
Indeterminate: The SoS behavior is dynamically unforeseen. \\
\hline
\end{tabular}
\label{tab:sos_comparison}
\end{center}
\end{table}

\begin{figure}[ht!]
    \centering
    \includegraphics[width=\linewidth]{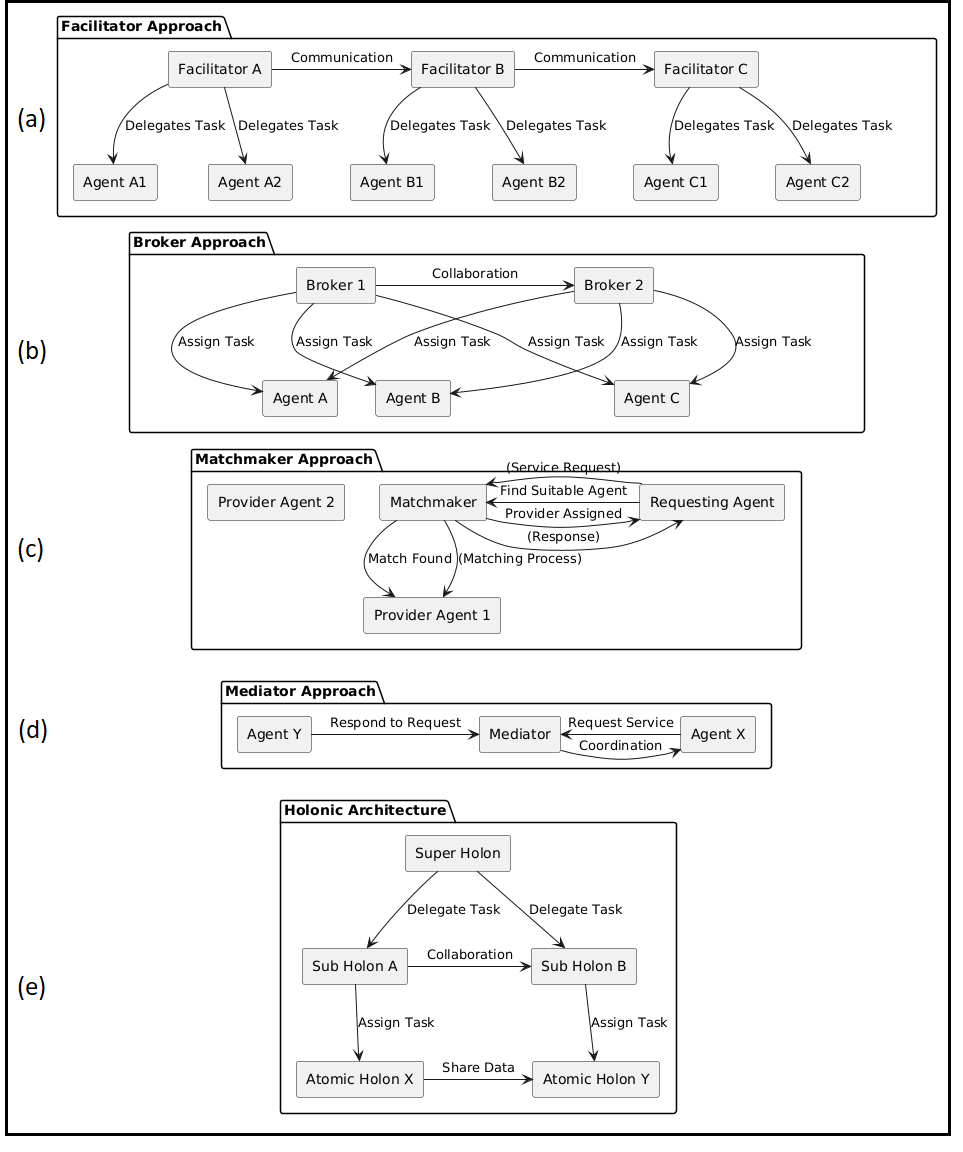}
    \caption{Comparison of different federated architectures, illustrating (a)~Facilitator, (b)~Broker, (c)~Matchmaking, (d)~Mediator, and (e)~Holonic  approaches.}
    \label{fig:federated_architectures}
\end{figure}

\section{Background}
\label{sec:background}

\subsection{Federated Architectures in Urban Air Mobility}
\label{sec:federated_architectures}

Traditional architectural approaches for UAM-SoS typically adopt either hierarchical or heterarchical models, each with inherent trade-offs. 
Hierarchical structures enable strong coordination but tend to be rigid, whereas heterarchical systems offer flexibility and fault tolerance but at the cost of increased complexity. 
Federated architectures present a balanced alternative, combining structured coordination with distributed autonomy, making them well-suited for large-scale, decentralized mobility ecosystems. 


Rather than relying on a centralized data repository, federated systems use message passing among agents and specialized \textit{middle agents} to facilitate coordination. This mechanism supports dynamic decision-making and real-time adaptation. As a result, federated architectures enhance fault tolerance, scalability, and operational flexibility---key attributes for UAM systems.

Several federated architectures have been proposed in \textit{multi-agent systems (MAS)} literature for managing large-scale, decentralized systems~\cite{sadik2019modeling}. The most relevant approaches for UAM are as follows (see Fig.~\ref{fig:federated_architectures}):

\subsubsection{Facilitator-based architecture} Utilizes specialized \textit{facilitator agents} to manage communication and task delegation among agents. Facilitators handle message routing and coordination, reducing communication overhead but introducing potential bottlenecks~\cite{benaskeur2008holonic}.

\subsection{Broker-based architecture}
Introduces \textit{broker agents} that mediate agent interactions. Unlike facilitators, brokers allow agents to communicate directly after the initial connection is established, improving efficiency but requiring robust service discovery protocols~\cite{guner2017message}.

\subsection{Matchmaking-based architecture}
Extends the broker model by allowing matched agents to establish \textit{direct, independent communication} without continuous broker involvement. This approach is widely used in service-oriented computing~\cite{lee2013matchtree}.

\subsection{Mediator-based architecture}
Expands on brokering and matchmaking by \textit{actively coordinating} interactions among dynamically formed clusters of agents. Mediators optimize collaboration by balancing workload and resolving conflicts~\cite{wiederhold1992mediators}.

\subsection{Holonic architecture}
Introduces a \textit{recursive hierarchy} of mediator-based coordination, forming \textit{dynamic holarchies} where high-level nodes focus on strategic goals while lower-level holons handle tactical operations. By combining \textit{distributed control} with \textit{hierarchical organization}, this approach is particularly well-suited to the dynamically changing demands of UAM, enabling resilient, goal-oriented decision-making while maintaining operational autonomy~\cite{sadik2016novel}.

However, these federated models lack an embedded reasoning capability: they cannot parse free-form requests or adapt plans dynamically.

\subsection{LLM-Enhanced Holonic Architecture}
\label{sec:llm-enhanced-holonic-architecture}

This architecture has two main components: the holon and specialized holons, which are explained below.

\subsubsection{Holon Structure}
\label{sec:holon_structure}

A holon is the fundamental unit of the proposed architecture, encapsulating a CS of the SoS. It is organized into three primary layers: reasoning, communication, and capabilities. These layers are described as follows:

The \textit{Reasoning Layer} delivers context-aware decision-making and dynamic task planning. An LLM serves as its intelligent engine, converting complex commands into actionable strategies while preserving context across system components. It parses user input into task specifications, enriches those specs with real-time data (e.g., weather, traffic) and domain knowledge, and then devises detailed plans---encoding them into the structured format needed by the Communication Layer.

The \textit{Communication Layer} bridges the Reasoning and Execution Layers, transforming LLM-generated strategies into executable commands. It implements instructions via middleware such as the Robot Operating System (ROS)~\cite{majumdar2017development}, monitors and allocates resources for task execution~\cite{xiang2024autonomous}, and routes real-time data among holons to ensure coordinated, system-wide operation.

In the \textit{Capabilities Layer}, each holon possesses task-specific capabilities, represented as integrated services.

\subsubsection{Specialized Holons for UAM Coordination}

The architecture incorporates several specialized holons, each designed to handle distinct functions within the SoS. Each holon's reasoning layer is fine-tuned for its role, providing tailored contextual processing for incoming commands.

The \textit{Supervisor Holon} orchestrates system-wide strategy and operations---aligning schedules, tracking resources, and coordinating inter-system tasks across the SoS.

The \textit{Planner Holon} takes the Supervisor Holons strategic goals and converts them into concrete task sequences.

The \textit{Task Holon} breaks down tasks into atomic actions and allocates execution responsibilities. It schedules and implements specific tasks, processes real-time environmental feedback, adapts to changing conditions, and reports status updates.

The \textit{Resource Holon} oversees both human and physical resources via two sub-holons. The \textit{Human Resource Holon} provides the human-machine interface. In contrast, the \textit{Machine Resource Holon} manages assets---vehicles, equipment, and facilities---optimizing their allocation based on task requirements and real-time conditions.

\section{LLM-Enhanced Holonic Solution for UAM}
\label{sec:case-study}

To demonstrate the practical application of our architecture in a real-world UAM context, we present a case study focusing on multimodal urban transportation. 
We apply the architecture to a conceptual scenario: a passenger uses electric scooters and air taxis in the UAM-SoS. This example highlights how the system orchestrates multimodal travel under UAM constraints.


\subsection{UAM-Specific Holon Implementation}

\begin{figure}[!t]
    \centering
    \includegraphics[width=\linewidth]{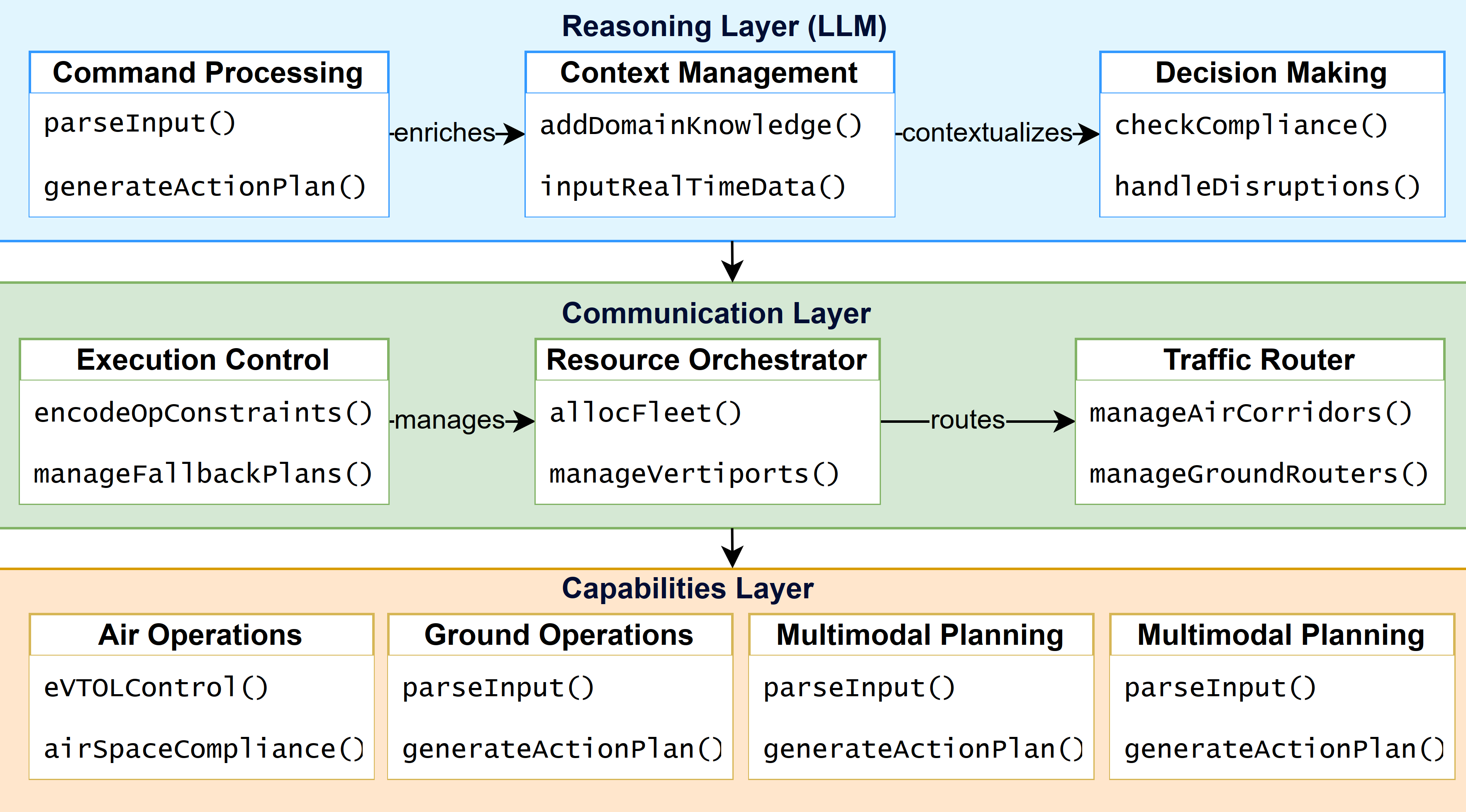}
    \caption{UAM-specific holon implementation showing the layers with specialized components for multimodal transportation coordination.}
    \label{fig:uam_holon_implementation}
\end{figure}

Fig.~\ref{fig:uam_holon_implementation} illustrates how the three-layer holon structure is tailored for UAM-SoS. In this implementation, the Reasoning Layer leverages LLMs for natural language processing of passenger requests and airspace compliance validation. 
The Communication Layer coordinates across diverse UAM assets, including eVTOL fleets, electric scooters, and vertiport systems. 
The Capabilities Layer encapsulates the operational status and real-time data from these UAM-specific resources. This tailored architecture enables the seamless integration of heterogeneous transportation modes while maintaining the autonomous and adaptive characteristics of the holonic approach.

\subsection{Specialized Holons for UAM}

\begin{figure}[!t]
    \centering
    \includegraphics[width=\linewidth]{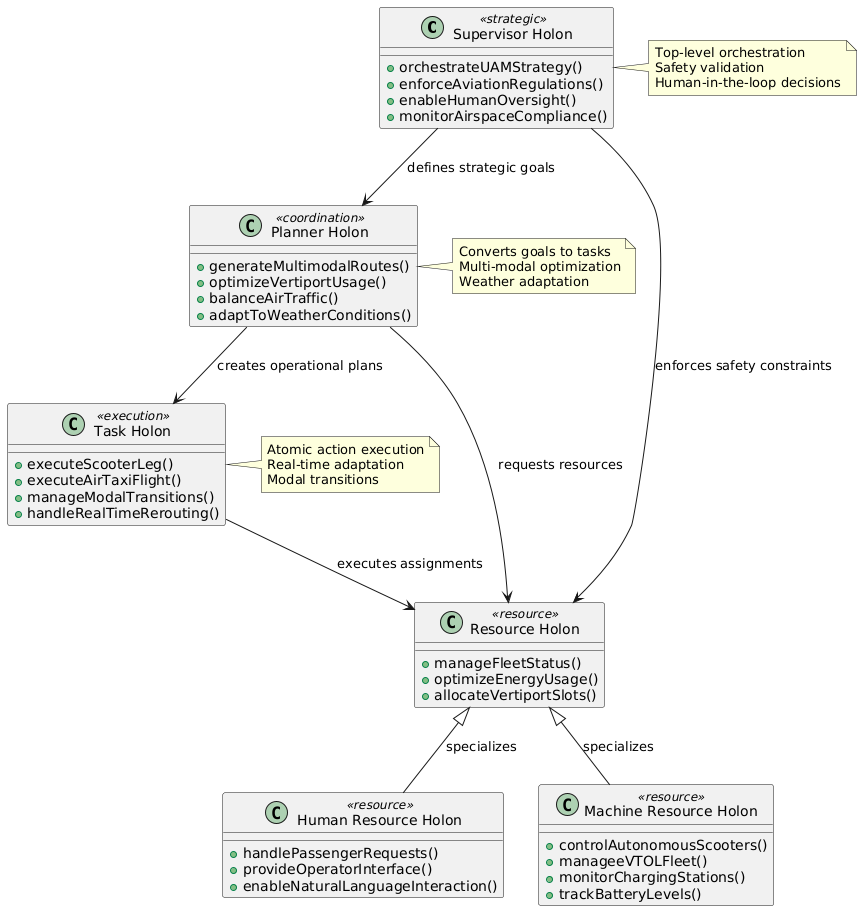}
    \caption{UAM Specialized holons and their workflow.}
    \label{fig:specialized_holons}
\end{figure}

In alignment with the principles discussed in Section~\ref{sec:llm-enhanced-holonic-architecture}, the transportation request is processed using the holonic architecture, where specialized holons contribute at different levels of decision-making and execution.
Fig.~\ref{fig:specialized_holons} illustrates these holons and their interactions.

\subsubsection{Supervisor Holon}

Acts as the top-level orchestrator and safety guard. It autonomously handles strategic tasks such as resource allocation, multimodal trip scheduling, and system-wide performance monitoring. Designed for human-in-the-loop operation, it provides supervisory interfaces through which human operators can review, approve, or override critical decisions, ensuring that safety-sensitive actions maintain human oversight. For safety-critical decisions (e.g., airspace clearance), it implements a three-step process:

\begin{itemize}
    \item LLM-generated plans are checked against aviation regulations (e.g., FAA rules).
    \item Real-time sensor feeds (e.g., collision-avoidance systems) confirm physical feasibility.
    \item Human operators review high-risk actions via a dashboard, with automated fallback plans triggered if approval is delayed.
\end{itemize}

\subsubsection{Plan Holon}
Takes Supervisor Holon's strategic goals and converts them into concrete task sequences. It selects appropriate transport modes, allocates resources, and optimizes their distribution across multiple UAM operational scenarios.

\subsubsection{Task Holon}
Breaks down each trip into sub-tasks (e.g., scooter ride, drone flight, and last-mile driving) and executes individual tasks such as a scooter leg or an air-taxi leg. It dynamically adapts to real-time conditions, monitors the local environment, responds to feedback loops, and provides continuous status updates.

\subsubsection{Resource Holons}
The Human Resource Holon interfaces with passengers and operators within the UAM ecosystem. The Machine Resource Holon controls assets---drones, air taxis, ground vehicles, and charging stations---dynamically optimizing their deployment to meet real-time UAM service requests.

\subsection{Decision-Making Enhancements}

Unlike classical planners or expert systems, which rely on explicit state-space models or handcrafted decision rules, our LLM-based Reasoning Layer learns patterns from large corpora to (1) generate semantically rich plans from ambiguous prompts and (2) revise those plans on the fly in response to new contextual inputs. 
This capability significantly improves adaptability by enabling real-time reasoning, autonomous plan validation, and proactive decision-making. Key enhancements for dynamic trip optimization include:

\begin{itemize}
    \item Intelligent Resource Matching: 
    The LLM considers context (traffic, battery levels) to pick the best scooter and air taxi for each leg.
    \item Dynamic Replanning: The LLM validates plans against real-time data (e.g., weather, airspace status) and triggers reroute (e.g., substituting ground taxis for air routes) during disruptions.
    \item Natural Language Coordination: 
    Passengers or operators can send updates in plain language (e.g. ‘I’m running late’), which the LLM uses to adjust schedules or priorities.
    
\end{itemize}

\subsection{Workflow and Execution Sequence}
\label{sec:workflow}

\begin{figure}[!t]
\centering
\includegraphics[width=\linewidth]{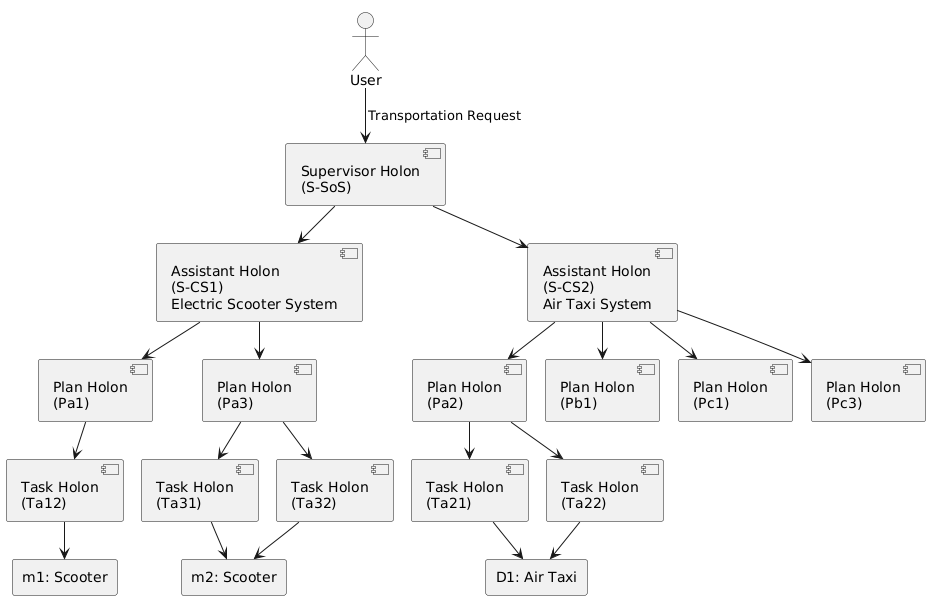}
\caption{Holonic architecture of the UAM-SoS, showing coordination among supervisor, assistant, plan, task, and resource holons for electric scooters and air taxis.}
\label{fig:3d-mobility-holonic-architecture}
\end{figure}

\begin{figure*}[!t]
\centering
\includegraphics[width=\linewidth]{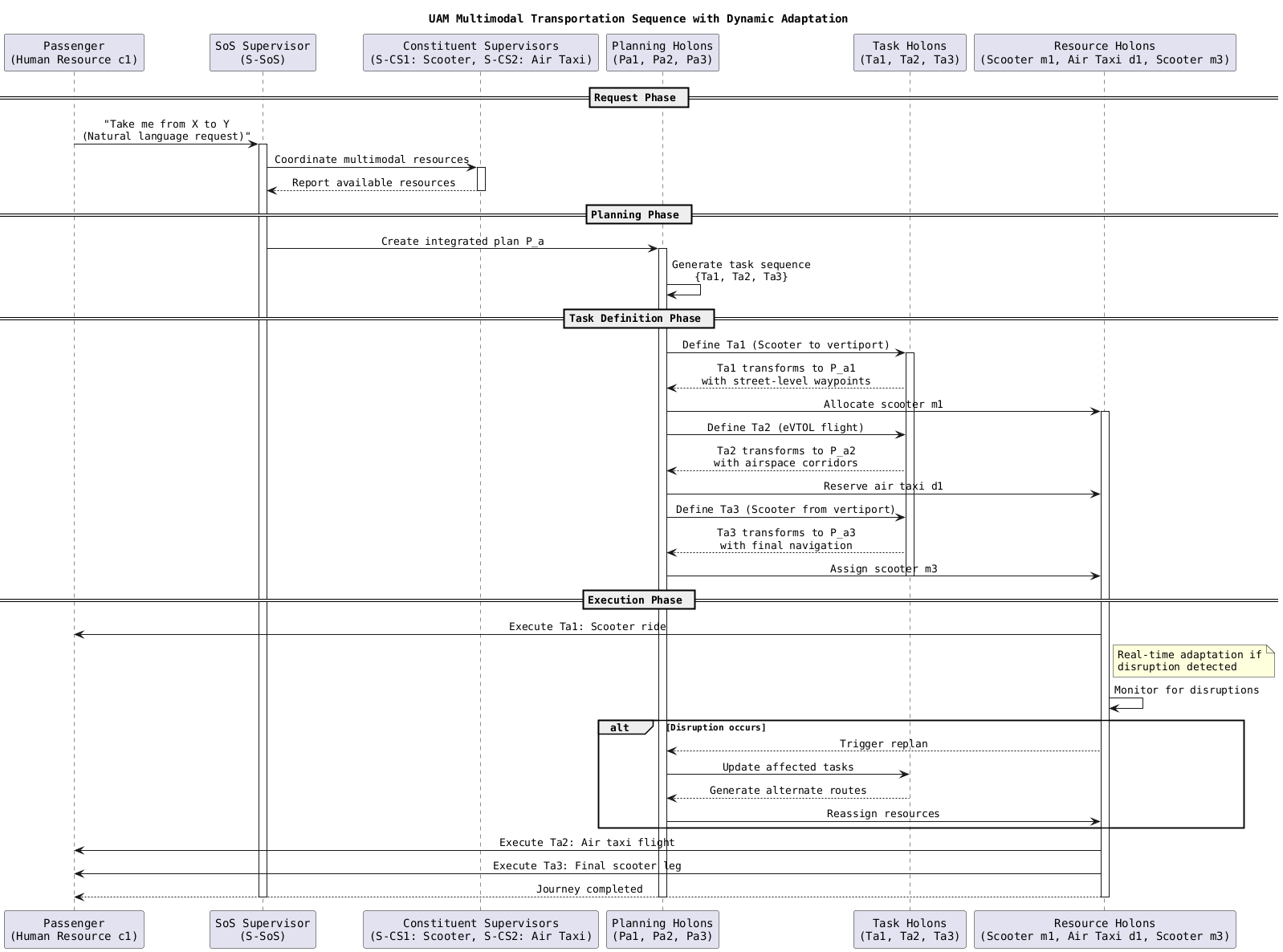}
\caption{Interaction sequence for an on-demand UAM flight.}
\label{fig:sequence-diagram}
\end{figure*}

The overall holonic architecture of the integrated UAM-SoS is shown in Fig.\ref{fig:3d-mobility-holonic-architecture}, while the detailed workflow is illustrated in the sequence diagram in Fig.\ref{fig:sequence-diagram}. The system operates as follows: A passenger (modeled as a Human Resource Holon $c_1$) requests a ride from point \emph{X} to point \emph{Y}.
The top‐level Supervisor Holon (\textit{S-SoS}) receives this request and collaborates with its two subordinate supervisors: Scooter Supervisor (\textit{S-CS1}), which manages a fleet of ground scooters, and AirTaxi Supervisor (\textit{S-CS2}), which oversees eVTOL flights.
Through their collaboration, the system constructs a multimodal journey plan.

The Planner Holon generates a trip structure $P_a = \{T_{a1}, T_{a2}, T_{a3}\}$ consisting of three coordinated legs: an electric scooter ride to the nearest vertiport ($T_{a1}$), an air-taxi flight to the urban core ($T_{a2}$), and a final scooter leg to the destination ($T_{a3}$). The planning process employs LLM-enhanced reasoning to refine each macroscopic leg into executable subtasks -- for instance, converting $T_{a1}$ into street-level navigation waypoints.

During execution, the holonic architecture enables autonomous adaptation: if disruptions occur (e.g., a blocked route during $T_{a1}$), the system automatically triggers replanning through real-time status monitoring and LLM-assisted reasoning. This closed-loop operation maintains service continuity without requiring manual intervention, demonstrating the system's resilience through dynamic resource coordination between ground and air transport modalities.

\section{Discussion}
\label{sec:discussion}

This case study demonstrates that the LLM-enhanced holonic architecture provides an intelligent, adaptive, and scalable solution for UAM-SoS, directly addressing the architectural requirements for SoS outlined in Table~\ref{tab:sos_comparison}.
Unlike traditional centralized models, the system achieves decentralized coordination through three UAM-critical capabilities:

\begin{enumerate}
    \item Decentralized Autonomy (Table~\ref{tab:sos_comparison}: \textit{Autonomy}): Each LLM-powered holon independently handles mission-critical tasks---translating passenger requests into airspace-compliant routes, negotiating landing slots with vertiports, and replanning for battery constraints---without central bottlenecks. This decentralization proves vital when coordinating hundreds of simultaneous UAM flights amid weather disruptions or priority vehicle rerouting.

    \item Context-Aware Airspace Adaptation (Table~\ref{tab:sos_comparison}: \textit{Emergence}): Holons continuously integrate live operational data via LLM prompts, including sudden no-fly zones, vertiport congestion alerts, or traffic pattern shifts. For example, a Scooter Holon can redirect to alternate landing sites after receiving a vertiport closure notice while maintaining passenger ETA guarantees through coordinated air corridor adjustments.

    \item Natural-Language Tasking (Table~\ref{tab:sos_comparison}: \textit{Interoperability}): 
    LLMs formalize plain-language inputs (e.g., "Avoid turbulence") into safety-compliant constraints, as demonstrated in Section~\ref{sec:workflow}.

\end{enumerate}

The recursive holarchy enables seamless scaling (Scalability, Table~\ref{tab:sos_comparison}), allowing new subsystems (e.g., vertiports) to self-integrate via LLM-driven discovery. 
The federated design avoids single-point failures (e.g., control tower outages), ensuring resilience (Security/Safety, Section~\ref{sec:introduction}).
This fusion of autonomy, adaptability, and interoperability enables our proposed solution to meet UAM's challenges while adhering to core SoS architectural principles.

\section{Conclusion}
\label{sec:conclusion}

In this research, we examined UAM challenges from a SoS perspective, emphasizing the decentralized, heterogeneous, and dynamic nature of UAM environments. We proposed a holonic architecture integrated with LLMs to enable modular, distributed control, and adaptive decision-making, addressing challenges in coordination, resource allocation, and operational safety.

The practical impact of our framework lies in its ability to transform UAM operations into a self-organizing, intelligent network. 
Integrating LLM reasoning with hierarchical holonic control enables higher throughput, more resilient fault tolerance, and enhanced safety.

Compared to traditional heuristic-based coordination strategies, our LLM-enhanced holonic model offers improved scalability, flexibility, and context awareness. However, deploying LLMs introduces computational overheads and latency trade-offs, which must be carefully managed in real-time UAM operations. Furthermore, the safety-critical nature of UAM demands robust approaches to data privacy, secure communication, and fault tolerance. Potential limitations also include regulatory challenges, hardware constraints, and ensuring data privacy compliance in decentralized systems.

Alternative AI paradigms, such as reinforcement learning (RL) and conventional multi-agent systems, offer complementary strengths. RL methods have shown promise in adaptive routing and resource allocation for air traffic management~\cite{lee2024advanced, paul2022graph}. The federated and graph-based multi-agent frameworks enable privacy-preserving, low-latency task allocation in UAM networks~\cite{wei2021scheduling}. However, they often rely on predefined protocols and lack the open-domain reasoning capabilities of LLM-enhanced holonic models.

Future work will focus on optimizing LLM inference pipelines to reduce computational costs, enhancing security mechanisms for safe inter-holon communication, and benchmarking our framework against conventional mobility coordination methods. We will also integrate bias-detection modules directly into the Reasoning Layer to ensure fairness and transparency in automated decision-making. Additionally, we aim to validate our approach through large-scale simulations, real-world testbeds, and collaboration with regulatory bodies to ensure compliance and operational readiness. Finally, we plan to explore hybrid AI frameworks that integrate reinforcement learning, federated learning, and LLM reasoning for enhanced adaptability and resilience in UAM-SoS deployments.

\bibliographystyle{IEEEtran}
\bibliography{main}

\begin{thebibliography}{10}
\providecommand{\url}[1]{#1}
\csname url@samestyle\endcsname
\providecommand{\newblock}{\relax}
\providecommand{\bibinfo}[2]{#2}
\providecommand{\BIBentrySTDinterwordspacing}{\spaceskip=0pt\relax}
\providecommand{\BIBentryALTinterwordstretchfactor}{4}
\providecommand{\BIBentryALTinterwordspacing}{\spaceskip=\fontdimen2\font plus
\BIBentryALTinterwordstretchfactor\fontdimen3\font minus \fontdimen4\font\relax}
\providecommand{\BIBforeignlanguage}[2]{{%
\expandafter\ifx\csname l@#1\endcsname\relax
\typeout{** WARNING: IEEEtran.bst: No hyphenation pattern has been}%
\typeout{** loaded for the language `#1'. Using the pattern for}%
\typeout{** the default language instead.}%
\else
\language=\csname l@#1\endcsname
\fi
#2}}
\providecommand{\BIBdecl}{\relax}
\BIBdecl

\bibitem{10388419}
H.~Wei, B.~Lou, Z.~Zhang, B.~Liang, F.-Y. Wang, and C.~Lv, ``Autonomous navigation for {eVTOL}: Review and future perspectives,'' \emph{IEEE Transactions on Intelligent Vehicles}, vol.~9, no.~2, pp. 4145--4171, 2024.

\bibitem{xiang2024autonomous}
S.~Xiang, A.~Xie, M.~Ye, X.~Yan, X.~Han, H.~Niu, Q.~Li, and H.~Huang, ``Autonomous {eVTOL}: A summary of researches and challenges,'' \emph{Green Energy and Intelligent Transportation}, vol.~3, no.~1, p. 100140, 2024.

\bibitem{koreaHerald}
\BIBentryALTinterwordspacing
T.~K. Herald, ``{Daily commute in greater Seoul takes 83 minutes: report},'' 2023, {Accessed: 2025-04-25}. [Online]. Available: \url{https://www.koreaherald.com/article/3285257}
\BIBentrySTDinterwordspacing

\bibitem{UNUrbanizaiton}
\BIBentryALTinterwordspacing
{United Nations}, ``{68\% of the world population projected to live in urban areas by 2050, says UN},'' 2018, {Accessed: 2025-04-25}. [Online]. Available: \url{https://www.un.org/development/desa/en/news/population/2018-revision-of-world-urbanization-prospects.html}
\BIBentrySTDinterwordspacing

\bibitem{moradi2024urban}
N.~Moradi, C.~Wang, and F.~Mafakheri, ``Urban air mobility for last-mile transportation: A review,'' \emph{Vehicles}, vol.~6, no.~3, pp. 1383--1414, 2024.

\bibitem{rubio2023urban}
C.~G. Rubio and T.~Rigaut, ``An urban air mobility system of systems {UAF} \& {MDAO} application case,'' in \emph{2023 18th Annual System of Systems Engineering Conference}.\hskip 1em plus 0.5em minus 0.4em\relax IEEE, 2023, pp. 01--08.

\bibitem{maier1998architecting}
M.~W. Maier, ``Architecting principles for systems-of-systems,'' \emph{Systems Engineering: The Journal of the International Council on Systems Engineering}, vol.~1, no.~4, pp. 267--284, 1998.

\bibitem{boardman2006system}
J.~Boardman and B.~Sauser, ``System of systems--the meaning of \textit{of},'' in \emph{2006 IEEE/SMC international conference on system of systems engineering}.\hskip 1em plus 0.5em minus 0.4em\relax IEEE, 2006, pp. 6--pp.

\bibitem{sadik2023self}
A.~R. Sadik, B.~Bolder, and P.~Subasic, ``A self-adaptive system of systems architecture to enable its ad-hoc scalability: unmanned vehicle fleet-mission control center case study,'' in \emph{Proceedings of the 7th International Conference on Intelligent Systems, Metaheuristics \& Swarm Intelligence}.\hskip 1em plus 0.5em minus 0.4em\relax ACM, 2023, pp. 111--118.

\bibitem{brulin2025system}
S.~Brulin and A.~Sadik, ``System design process for multi-layered transport networks,'' Jan.~30 2025, uS Patent App. 18/360,790.

\bibitem{ISO2019}
\emph{Systems and software engineering — System of systems (SoS) considerations in life cycle stages}, ISO/IEC/IEEE 21839 Std., 2019.

\bibitem{fontaine2023urban}
P.~Fontaine, ``{Urban air mobility (UAM) concept of operations},'' \emph{Federal Aviation Administration}, vol. 800, 2023.

\bibitem{sengupta2023urban}
D.~Sengupta and S.~K. Das, ``Urban air mobility: Vision, challenges and opportunities,'' in \emph{2023 IEEE 24th International Conference on High Performance Switching and Routing (HPSR)}.\hskip 1em plus 0.5em minus 0.4em\relax IEEE, 2023, pp. 1--6.

\bibitem{Henshaw2013}
M.~J. Henshaw, ``Systems of systems, cyber-physical systems, and the internet of things,'' in \emph{INCOSE International Symposium}, vol.~23, no.~1.\hskip 1em plus 0.5em minus 0.4em\relax Wiley Online Library, 2013, pp. 55--68.

\bibitem{DANSE}
G.~A. Lewis, S.~A. Lavenberg, J.~A. Keenan, and D.~B. Smith, ``An approach to analyzing system-of-systems interoperability,'' in \emph{Proceedings of the 21st ACM International Conference on Software and Systems Process}.\hskip 1em plus 0.5em minus 0.4em\relax ACM, 2016, pp. 70--79.

\bibitem{elhabbash2024principled}
A.~Elhabbash, Y.~Elkhatib, V.~Nundloll, V.~S. Marco, and G.~S. Blair, ``Principled and automated system of systems composition using an ontological architecture,'' \emph{Future Generation Computer Systems}, vol. 157, pp. 499--515, 2024.

\bibitem{blair2015holons}
G.~S. Blair, Y.~Bromberg, G.~Coulson, Y.~Elkhatib, L.~R{\'{e}}veill{\`{e}}re, H.~B. Ribeiro, E.~Rivi{\`{e}}re, and F.~Ta{\"{\i}}ani, ``Holons: Towards a systematic approach to composing systems of systems,'' in \emph{Proceedings of the 14th International Workshop on Adaptive and Reflective Middleware}.\hskip 1em plus 0.5em minus 0.4em\relax {ACM}, 2015, pp. 5:1--5:6.

\bibitem{indriago2016h2cm}
C.~Indriago, O.~Cardin, N.~Rakoto, P.~Castagna, and E.~Chac{\`o}n, ``{H2CM}: A holonic architecture for flexible hybrid control systems,'' \emph{Computers in industry}, vol.~77, pp. 15--28, 2016.

\bibitem{alahakoon2023self}
D.~Alahakoon, R.~Nawaratne, Y.~Xu, D.~De~Silva, U.~Sivarajah, and B.~Gupta, ``Self-building artificial intelligence and machine learning to empower big data analytics in smart cities,'' \emph{Information Systems Frontiers}, pp. 1--20, 2023.

\bibitem{Sha2024Generative}
Z.~Sha, W.~Yue, S.~Wang, N.~Cheng, J.~Wu, and C.~Li, ``Generative {AI}-enabled sensing and communication integration for urban air mobility,'' \emph{2024 IEEE 99th Vehicular Technology Conference}, pp. 1--5, 2024.

\bibitem{ashfaq2025llm}
M.~Ashfaq, A.~R. Sadik, T.~Mikkonen, M.~Waseem, and N.~M{\"a}kitalo, ``{LLM}-ehnanced holonic architecture for ad-hoc scalable {SoS},'' \emph{arXiv preprint arXiv:2501.07992}, 2025.

\bibitem{sadik2019modeling}
A.~R. Sadik, C.~Goerick, and M.~Muehlig, ``Modeling and simulation of a multi-robot system architecture,'' in \emph{2019 International Conference on Mechatronics, Robotics and Systems Engineering}.\hskip 1em plus 0.5em minus 0.4em\relax IEEE, 2019, pp. 8--14.

\bibitem{benaskeur2008holonic}
A.~R. Benaskeur and H.~Irandoust, ``Holonic approach for control and coordination of distributed sensors,'' Defence R\&D Canada--Valcartier, Quebec, Canada, Technical Report DRDC--Valcartier TR 2008-015, Aug. 2008.

\bibitem{guner2017message}
A.~Guner, K.~Kurtel, and U.~Celikkan, ``A message broker based architecture for context aware {IoT} application development,'' in \emph{2017 International Conference on Computer Science and Engineering (UBMK)}.\hskip 1em plus 0.5em minus 0.4em\relax IEEE, 2017, pp. 233--238.

\bibitem{lee2013matchtree}
K.~Lee, T.~Choi, P.~O. Boykin, and R.~J. Figueiredo, ``Matchtree: Flexible, scalable, and fault-tolerant wide-area resource discovery with distributed matchmaking and aggregation,'' \emph{Future Generation Computer Systems}, vol.~29, no.~6, pp. 1596--1610, 2013.

\bibitem{wiederhold1992mediators}
G.~Wiederhold, ``Mediators in the architecture of future information systems,'' \emph{Computer}, vol.~25, no.~3, pp. 38--49, 1992.

\bibitem{sadik2016novel}
A.~R. Sadik and B.~Urban, ``A novel implementation approach for resource holons in reconfigurable product manufacturing cell,'' in \emph{International Conference on Informatics in Control, Automation and Robotics}, vol.~2.\hskip 1em plus 0.5em minus 0.4em\relax SCITEPRESS, 2016, pp. 130--139.

\bibitem{majumdar2017development}
A.~Majumdar, N.~Gamez, P.~Benavidez, and M.~Jamshidi, ``Development of robot operating system {(ROS)} compatible open source quadcopter flight controller and interface,'' in \emph{2017 12th System of Systems Engineering Conference (SoSE)}.\hskip 1em plus 0.5em minus 0.4em\relax IEEE, 2017, pp. 1--6.

\bibitem{lee2024advanced}
S.~Lee, G.~S. Kim, S.~Park, and J.~Kim, ``Advanced taxiing path guidance using multi-agent reinforcement learning for air traffic management,'' in \emph{2024 22nd International Symposium on Modeling and Optimization in Mobile, Ad Hoc, and Wireless Networks (WiOpt)}.\hskip 1em plus 0.5em minus 0.4em\relax IEEE, 2024, pp. 305--312.

\bibitem{paul2022graph}
S.~Paul and S.~Chowdhury, ``A graph-based reinforcement learning framework for urban air mobility fleet scheduling,'' in \emph{AIAA AVIATION 2022 Forum}, 2022, p. 3911.

\bibitem{wei2021scheduling}
Q.~Wei, G.~Nilsson, and S.~Coogan, ``Scheduling of urban air mobility services with limited landing capacity and uncertain travel times,'' in \emph{2021 American Control Conference (ACC)}.\hskip 1em plus 0.5em minus 0.4em\relax IEEE, 2021, pp. 1681--1686.

\end{thebibliography}

\end{document}